\documentclass{article}

\usepackage[utf8]{inputenc}
\usepackage[T1]{fontenc}
\usepackage{amsmath, amssymb, amsfonts}
\usepackage{graphicx}
\usepackage{hyperref}
\usepackage{booktabs}
\usepackage{float}
\usepackage{caption}
\usepackage{subcaption}
\usepackage{url}
\usepackage{color}
\usepackage{geometry}
\usepackage[numbers]{natbib}

\title{Understanding Hidden Computations in Chain-of-Thought Reasoning}

\author{Aryasomayajula Ram Bharadwaj \\
Independent Researcher \\
\texttt{ram.bharadwaj.arya@gmail.com}}

\date{}

\begin{document}

\maketitle

\begin{abstract}
Chain-of-Thought (CoT) prompting has significantly enhanced the reasoning abilities of large language models. However, recent studies have shown that models can still perform complex reasoning tasks even when the CoT is replaced with filler(hidden) characters (e.g., ``\ldots''), leaving open questions about how models internally process and represent reasoning steps. In this paper, we investigate methods to decode these hidden characters in transformer models trained with filler CoT sequences. By analyzing layer-wise representations using the logit lens method and examining token rankings, we demonstrate that the hidden characters can be recovered without loss of performance. Our findings provide insights into the internal mechanisms of transformer models and open avenues for improving interpretability and transparency in language model reasoning.
\end{abstract}

\section{Introduction}

Chain-of-Thought (CoT) prompting has emerged as a powerful technique for improving the performance of large language models (LLMs) on complex reasoning tasks \cite{wei2022chain}. By encouraging models to generate intermediate reasoning steps, CoT prompting enables models to tackle problems that require multi-step reasoning. However, recent work by \cite{pfau2023let} demonstrates that these improvements can be achieved even when the CoT is replaced with hidden or filler characters (e.g., ``\ldots''). This raises intriguing questions about the nature of the computations being performed within these models and how they internally represent reasoning steps when the explicit chain of thought is obscured.

In this paper, we build upon the findings of \cite{pfau2023let} and investigate methods to decode these hidden characters in transformer models trained with filler CoT sequences. Focusing on the 3SUM task as a case study, we analyze the internal representations of the model using the logit lens method \cite{nostalgebraist2020} and examine token rankings during the decoding process. Our goal is to understand how the model processes and retains information when the explicit reasoning steps are replaced with filler tokens, and whether the hidden characters can be recovered without loss of performance.

Our contributions are as follows:

\begin{itemize}
    \item We provide a detailed analysis of the layer-wise representations in a transformer model trained with filler CoT sequences, revealing how the hidden characters evolve across layers.
    \item We demonstrate that the hidden characters can be recovered by examining higher-ranked tokens during decoding, without compromising the model's performance on the task.
    \item We discuss the implications of our findings for model interpretability and suggest directions for future research in understanding hidden computations involving filler characters in language models.
\end{itemize}

\newpage
\section{Background}

In this section, we provide an overview of the key concepts and previous work relevant to our study, including Chain-of-Thought prompting, hidden chain-of-thought with filler tokens, the 3SUM task, and the logit lens method.

\subsection{Chain-of-Thought Prompting}

Chain-of-Thought (CoT) prompting involves providing language models with prompts that include intermediate reasoning steps leading to the final answer \cite{wei2022chain}. This technique has been shown to improve the performance of large language models on tasks requiring multi-step reasoning, such as mathematical problem-solving and commonsense reasoning.

Formally, given a question $Q$, the model is prompted to generate a sequence of tokens that includes a chain of reasoning $R$ followed by the final answer $A$. The desired output is thus $S = R \circ A$, where $\circ$ denotes concatenation. For example, for the arithmetic question ``What is the result of (12 plus 15) multiplied by 2 minus 5?'', the chain of thought could be ``First, add 12 and 15 to get 27. Next, multiply 27 by 2 to obtain 54. Then, subtract 5 from 54 to arrive at 49'', leading to the final answer ``49''.

\subsection{Hidden Chain-of-Thought with Filler Tokens}

In a variant of CoT prompting, the intermediate reasoning steps are replaced with filler characters (e.g., ``\ldots''), resulting in a hidden chain of thought. That is, instead of outputting $S = R \circ A$, the model outputs $S = F \circ A$, where $F$ is a sequence of filler tokens. \cite{pfau2023let} found that models trained with such filler CoT sequences can still perform well on reasoning tasks, suggesting that meaningful computation occurs internally despite the absence of explicit reasoning steps in the output.

This phenomenon raises questions about how models internally process and represent reasoning steps when the chain of thought is obscured. Understanding this internal processing is important for model interpretability and could have implications for model safety and controllability.

\subsection{The 3SUM Task}

The 3SUM task involves determining whether any three numbers in a given list sum to zero. Formally, given a list of integers $S = \{s_1, s_2, \dots, s_n\}$, the task is to decide whether there exist indices $i, j, k$ (with $i \ne j \ne k$) such that $s_i + s_j + s_k = 0$.

The 3SUM problem is a well-known computational problem and serves as a proxy for more complex reasoning tasks in the context of language models \cite{pfau2023let}. By using a mathematical problem that requires combinatorial reasoning, we can study the model's ability to perform internal computations and understand how it processes the problem when the chain of thought is hidden.

\subsection{Logit Lens Method}

The logit lens method, introduced by \cite{nostalgebraist2020}, provides a way to inspect the internal representations of a language model by mapping the activations at each layer back to the vocabulary space. Specifically, at each layer $l$, the hidden state $\mathbf{h}^l$ is projected onto the vocabulary logits using the output embedding matrix $\mathbf{W}_{\text{out}}$, yielding $\mathbf{z}^l = \mathbf{h}^l \mathbf{W}_{\text{out}}$.

By applying the softmax function, we obtain a probability distribution over the vocabulary at each layer. This method allows us to observe the model's intermediate predictions and gain insights into how information is processed and transformed across layers. By examining the top predicted tokens at each layer, we can infer the model's evolving ``thoughts'' as it processes the input and generates the output.

\newpage
\section{Related Work}

Understanding the internal mechanisms of Chain-of-Thought (CoT) reasoning has been a significant focus of recent research. \cite{pfau2023let} demonstrated that models can perform reasoning tasks even when explicit CoT is replaced with filler tokens, suggesting the presence of hidden computations. This finding challenges the assumption that visible CoT directly reflects the model's reasoning process and indicates that models may retain and utilize internal reasoning pathways that are not immediately observable in the output.

Further studies have explored the faithfulness and reliability of CoT reasoning. Lanham et al. \cite{lanham2023measuringfaithfulnesschainofthoughtreasoning} investigated the faithfulness of CoT explanations, revealing that as models become larger and more capable, the faithfulness of their reasoning often decreases across various tasks. Their work highlights that CoT's performance improvements do not solely stem from added computational steps or specific phrasing, but also raise concerns about the genuine alignment between stated reasoning and actual cognitive processes.

Turpin et al. \cite{turpin2023languagemodelsdontsay} found that CoT explanations can systematically misrepresent the true reasons behind a model's predictions. They demonstrated that models might generate plausible yet misleading explanations, especially when influenced by biased inputs, leading to significant drops in accuracy and reinforcing stereotypes without acknowledging underlying biases.

Addressing these challenges, Radhakrishnan et al. \cite{radhakrishnan2023questiondecompositionimprovesfaithfulness} proposed decomposing questions into simpler sub-questions to enhance the faithfulness of model-generated reasoning. Their decomposition-based methods not only improve the reliability of CoT explanations but also maintain performance gains, suggesting a viable pathway to more transparent and verifiable reasoning processes in large language models.

Collectively, these studies underscore the complexity of CoT reasoning and the necessity for ongoing efforts to ensure that model-generated explanations are both faithful and transparent.

\section{Methodology}

In this section, we describe the experimental setup, including the model architecture, training procedure, and the analysis methods used to investigate the hidden computations.

\subsection{Model and Training}

We employed a transformer language model based on the LLaMA architecture \cite{touvron2023llama}, with 4 layers, a hidden dimension of 384, and 6 attention heads, totaling 34 million parameters. The model was randomly initialized and trained from scratch.

\newpage
\section{Experimental Setup}

\subsection{Dataset Generation}

We generated a synthetic dataset for the \textbf{Match-3} task, an extension of the classic 3SUM problem. Each input instance comprises a sequence of integer tuples, where each integer within a tuple is sampled uniformly from the range $[0, 9]$. Specifically, each input sequence \( S = \{s_1, s_2, \dots, s_7\} \) consists of 7 tuples, each of dimension 3. The corresponding labels indicate whether any triplet of tuples within the sequence sums to zero modulo 10, thereby ensuring a balanced distribution of positive and negative examples.

\subsubsection{True and Corrupted Instances}

To create this dataset, we defined two types of instances:

\begin{itemize}
    \item \textbf{True Instances:} Sequences where at least one triplet of tuples satisfies the 3SUM condition modulo 10. The generation process ensures uniform sampling of tuples and incorporates symmetry to maintain balanced labels.
    
    \item \textbf{Corrupted Instances:} Sequences with intentional perturbations based on a specified corruption rate (set to \( \frac{4}{3} \) in our experiments). Corruptions involve altering tuple values to disrupt existing valid triplets or to introduce new ones. This approach helps in creating challenging examples that test the model's robustness.
\end{itemize}

\subsubsection{Motivation for Corrupted Instances}

Corrupted instances are introduced to evaluate the model's ability to generalize and maintain performance in the presence of noise or alterations in the input data. By disrupting valid triplets or creating new ones, we assess whether the model can still accurately determine the presence of a 3SUM condition, thereby testing its internal reasoning capabilities.

\subsection{Instance-Adaptive Chain of Thought}

Instance-Adaptive Chain-of-Thought (CoT) differs from parallelizable CoT in that it requires caching sub-problem solutions within token outputs. Specifically, in instance-adaptive computation, the operations performed in later CoT tokens depend on the results obtained from earlier CoT tokens. This dependency structure is incompatible with the parallel processing nature of filler token computation.

In the context of the 3SUM problem, our instance-adaptive CoT approach decomposes the problem into dimension-wise 3SUMs. For each triplet of tuples, a dimension-wise summation is computed only if the previous dimension's summation equals zero. This creates an instance-adaptive dependency where the computation of one dimension influences the subsequent computations. For example, if the sum of the first dimension is zero, the model proceeds to compute the sum of the second dimension; otherwise, it skips to the next triplet. This sequential dependency ensures that the CoT reflects the logical progression of solving the problem step-by-step.

\subsection{Data Generation Procedure}

The data generation process for both parallelizable and instance-adaptive CoT follows identical input sampling procedures. The Chain-of-Thought generation is implemented as follows:

Given an input sequence, for example, ``A15 B75 C22 D13'', the corresponding CoT is generated as ``: A B C 15 75 22 2 B C D 75 22 13 0 ANS True''. The generation process for each triplet, such as ``A B C'' and ``B C D'', involves the following steps:

\begin{enumerate}
    \item \textbf{Triple Listing:} For each triplet, if the sum of the first dimension is zero, list the individual triple (e.g., ``: A B C'').
    \item \textbf{Value Listing:} List the values of the triples by copying from the input sequence (e.g., ``15 75 22'').
    \item \textbf{Summation Result:} Compute and list the result of summing the given triple in each dimension modulo 10 (e.g., ``2'' since \( (15 + 75 + 22) \mod 10 = 2 \) for the second dimension).
\end{enumerate}

In our example, the triplet ``A B C'' sums to 0 in the first dimension but sums to 2 in the second dimension. Conversely, the triplet ``B C D'' sums to 0 in both dimensions, thereby satisfying the 3SUM condition for this input. The CoT annotations provide intermediate reasoning steps that trace the computation of these sums, facilitating better interpretability and debugging of the model's decision-making process.

The generation process also includes options for adding fillers and controlling the inclusion of CoT annotations, ensuring a diverse and comprehensive dataset. Specifically, the dataset includes mixtures of filler-token sequences and instance-adaptive sequences.

\subsubsection{Dataset Parameters}

The dataset generation parameters were set as follows:

\begin{itemize}
    \item \textbf{Training Samples:} 10,000,000 instances
    \item \textbf{Testing Samples:} 2,000 instances
    \item \textbf{Tuple Dimension:} 3
    \item \textbf{Modulus (\( \text{mod} \)):} 10
    \item \textbf{Sequence Length:} 7
    \item \textbf{True Instance Rate:} 50\%
    \item \textbf{CoT Rate:} 50\%
    \item \textbf{No Filler Rate:} 0\%
    \item \textbf{Corruption Rate:} \( \frac{4}{3} \)
\end{itemize}

Inputs are uniformly sampled within the specified range, and the generation process incorporates both true and corrupted instances to balance the dataset. Corrupted instances are generated by introducing perturbations based on the corruption rate, which alters tuple values to either disrupt existing valid triplets or create new ones. This strategy enhances the dataset's diversity and challenges the model to generalize effectively.

\subsection{Hyperparameter Configuration}

The model was trained using the following hyperparameters:

\begin{itemize}
    \item \textbf{Optimizer:} Adam
    \item \textbf{Learning Rate:} \(1 \times 10^{-4}\)
    \item \textbf{Batch Size:} 256
    \item \textbf{Number of Epochs:} 5
\end{itemize}

These settings were chosen to balance training efficiency and model performance, ensuring adequate learning without overfitting.

\subsection{Layer-wise Representation Analysis}

To investigate how the model processes the hidden characters across layers, we employed the logit lens method \cite{nostalgebraist2020}. At each layer $l$, we extracted the hidden states $\mathbf{h}^l$ and projected them onto the vocabulary logits using the output embedding matrix $\mathbf{W}_{\text{out}}$, yielding $\mathbf{z}^l = \mathbf{h}^l \mathbf{W}_{\text{out}}$.

By applying the softmax function to $\mathbf{z}^l$, we obtained the probability distribution over the vocabulary at each layer. This allowed us to examine the top predicted tokens at each layer and observe how the model's internal representations evolve across layers. Specifically, we were interested in whether the filler tokens or the original reasoning steps were predominant in the predictions at different layers.

\subsection{Token Ranking Analysis}

Building upon the observations from the layer-wise analysis, we conducted a token ranking analysis during the decoding process. For each position in the output sequence, we examined the top $k$ candidate tokens based on their predicted probabilities. Our aim was to determine whether the original, non-filler CoT tokens appeared among the lower-ranked candidates when the filler token was the top prediction.

By analyzing the token rankings, we assessed whether the model retains the hidden characters beneath the filler tokens and whether these computations can be recovered by considering lower-ranked tokens. This analysis provides insights into the model's internal processing and the extent to which the hidden characters are accessible.

\subsection{Modified Decoding Algorithm}
\label{ssec:Filler tokens}

Based on the insights from the token ranking analysis, we implemented a modified greedy autoregressive decoding algorithm to recover the hidden characters. The algorithm operates as follows:

\begin{enumerate}
    \item Initialize the output sequence with the start token.
    \item For each decoding step $t$:
    \begin{enumerate}
        \item Compute the probability distribution over the vocabulary using the current hidden state.
        \item If the top-ranked token is the filler token, select the highest-ranked non-filler token as the output for position $t$.
        \item Otherwise, select the top-ranked token as usual.
        \item Update the hidden state based on the selected token.
    \end{enumerate}
    \item Continue the process until the end-of-sequence token is generated or the maximum sequence length is reached.
\end{enumerate}

This modified decoding algorithm allows us to recover the hidden characters by bypassing the filler tokens when they are the top prediction. By selecting the next most probable non-filler token, we can reconstruct the original reasoning steps without compromising the model's performance on the task.

\section{Results and Discussion}

In this section, we present the results of our experiments and discuss their implications for understanding hidden characters in transformer models.

\subsection{Layer-wise Representation Analysis}

Our layer-wise analysis revealed a gradual evolution of representations across the model's layers. In the initial layers (layers 1 and 2), the model's activations corresponded to the raw numerical sequences associated with the 3SUM problem's chain of thought. The top predicted tokens at these layers were primarily numerical tokens representing elements of the input sequence and intermediate calculations.

Starting from layer 3, we observed the emergence of filler tokens among the top-ranked predictions. The filler token (``\ldots'') began to appear more frequently as the top prediction, indicating that the model was starting to shift its focus toward producing the expected output format with filler tokens.

By layer 4 (the final layer), the filler token dominated the top predictions, and the original numerical tokens were relegated to lower ranks (rank-2). This pattern suggests that the model performs the necessary computations in the earlier layers and then overwrites the intermediate representations with filler tokens in the later layers to produce the expected output.

\begin{figure}[H]
\centering
\includegraphics[width=0.8\textwidth]{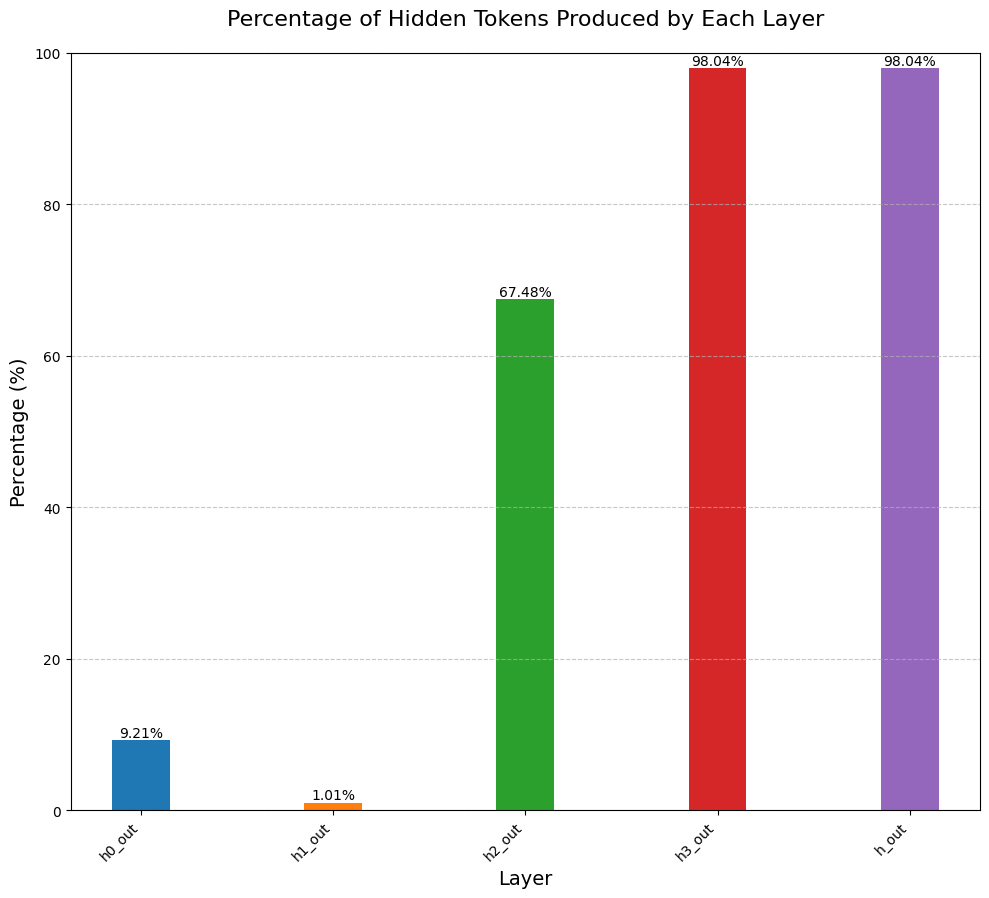}
\caption{Percentage of filler tokens among top predictions across layers}
\label{fig:hidden_token_percentages}
\end{figure}

Figure~\ref{fig:hidden_token_percentages} illustrates the percentage of filler tokens among the top predictions at each layer. The transition from numerical tokens to filler tokens across layers highlights how the model balances internal computation with output formatting.

\subsection{Token Ranking Analysis}

In the token ranking analysis, we found that while the filler token was consistently the top-ranked token at each decoding step in the later layers, the original, non-filler CoT tokens remained among the lower-ranked candidates. Specifically, the second-ranked token often corresponded to the numerical tokens representing the hidden reasoning steps.

This finding supports the hypothesis that the model retains the hidden characters beneath the filler tokens. The presence of the original reasoning tokens among the lower-ranked predictions indicates that the model internally processes the reasoning steps but prioritizes the filler tokens in the output to match the training targets.

\subsection{Decoding Methods Comparison}

We compared our modified decoding algorithm with two baselines:

\begin{itemize}
    \item \textbf{Standard Greedy Decoding:} Outputs the filler tokens as in the original training data.
    \item \textbf{Random Token Replacement:} Replaces filler tokens with randomly selected tokens from the vocabulary.
\end{itemize}

\begin{figure}[H]
\centering
\includegraphics[width=0.8\textwidth]{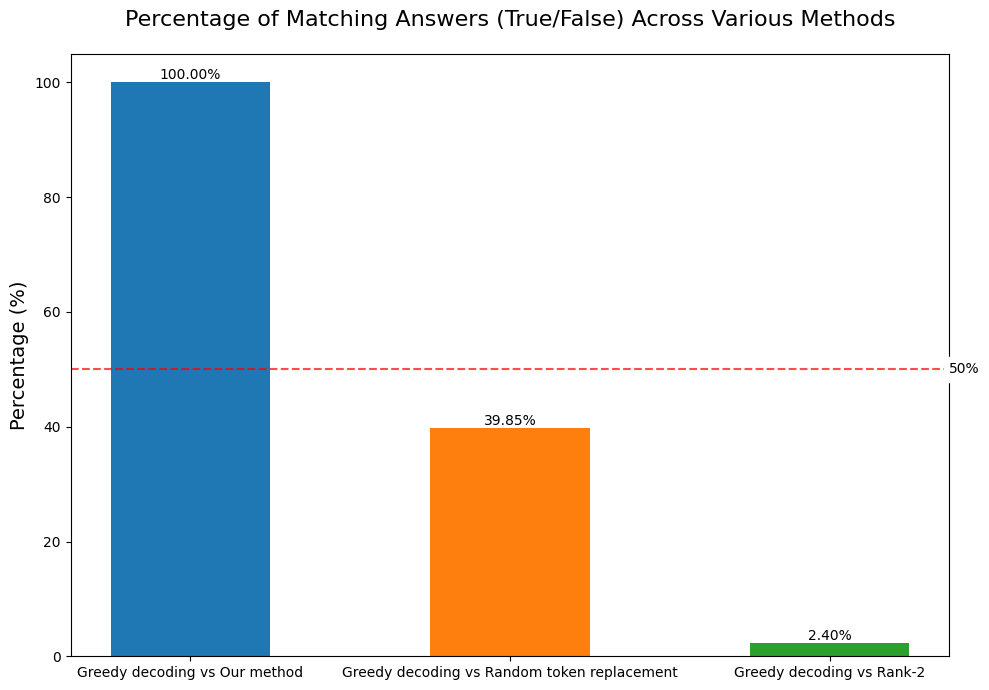}
\caption{Comparison of decoding methods: Our method achieves higher accuracy in recovering hidden characters compared to random token replacement}
\label{fig:token_comparison}
\end{figure}

Figure~\ref{fig:token_comparison} shows the performance of these methods in recovering the hidden characters.

Our method significantly outperformed the random replacement baseline and provided meaningful reasoning steps aligned with the model's internal computations. The standard greedy decoding, while aligning with training targets, does not reveal the hidden characters, highlighting the effectiveness of our approach in recovering internal reasoning.

\subsection{Discussion}

Our findings indicate that the model internally performs the necessary computations for the 3SUM task and that these computations can be recovered by examining lower-ranked tokens during decoding. The model appears to perform the reasoning in the earlier layers and then overwrites the intermediate representations with filler tokens in the later layers to produce the expected output.

This overwriting behavior may involve mechanisms such as induction heads \cite{elhage2021mathematical}, where the model learns to copy or overwrite tokens based on patterns in the data. Understanding how the model manages the trade-off between performing computations and producing the expected output could provide valuable insights into the internal workings of transformer models.

Our analysis contributes to the broader understanding of model interpretability and highlights the importance of examining internal representations to gain insights into how models process and represent information. By uncovering the hidden characters, we can develop methods to better control and interpret the outputs of language models.

\textbf{Limitations:} One limitation of our study is that it focuses on a synthetic task (3SUM) and a relatively small transformer model. While this allows for controlled experimentation, it may not fully capture the complexities of real-world language tasks and larger models. Future work should explore whether similar phenomena occur in larger models and more complex tasks.

\newpage
\section{Conclusion}

We have presented an analysis of computations involving filler/hidden characters in transformer models trained with filler CoT sequences on the 3SUM task. By utilizing the logit lens method and examining token rankings, we demonstrated that the hidden characters can be recovered without loss of performance. Our findings shed light on how models internally process and represent reasoning steps when the explicit chain of thought is obscured.

This work contributes to the broader understanding of model interpretability and opens avenues for improving transparency in language model reasoning. By uncovering the hidden computations, we can gain insights into the internal mechanisms of chain of thought reasoning in language models.

\section{Future Work}

Future research should focus on exploring whether specific circuits, such as induction heads or particular attention patterns, are involved could enhance our understanding of how models balance computation and output formatting.

Moreover, applying similar analysis techniques to other tasks and models could assess the generality of our findings. Extending the investigation to natural language tasks and larger models would help determine whether the observed phenomena persist in more complex settings.
\vfill
The code used for the experiments and analysis in this paper is available on GitHub \href{https://github.com/rokosbasilisk/filler_tokens}{here}.
\newpage
\appendix

\section{Layer-wise View of Sequences Generated via Various Decoding Methods}

In this appendix, we provide visualizations of the sequences generated using different decoding methods to illustrate how the hidden characters can be recovered.

\begin{figure}[H]
    \centering
    \includegraphics[width=\textwidth]{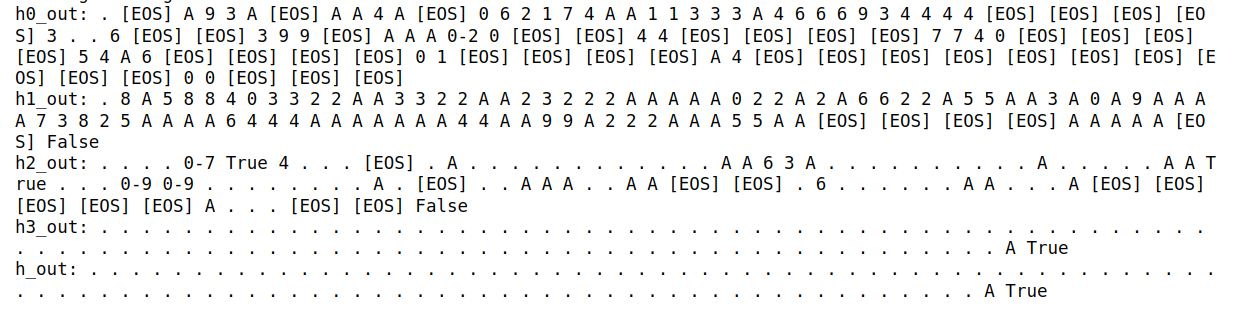}
    \caption{Greedy Decoding: The model outputs filler tokens followed by the final answer}
    \label{fig:greedy}
\end{figure}

\begin{figure}[H]
    \centering
    \includegraphics[width=\textwidth]{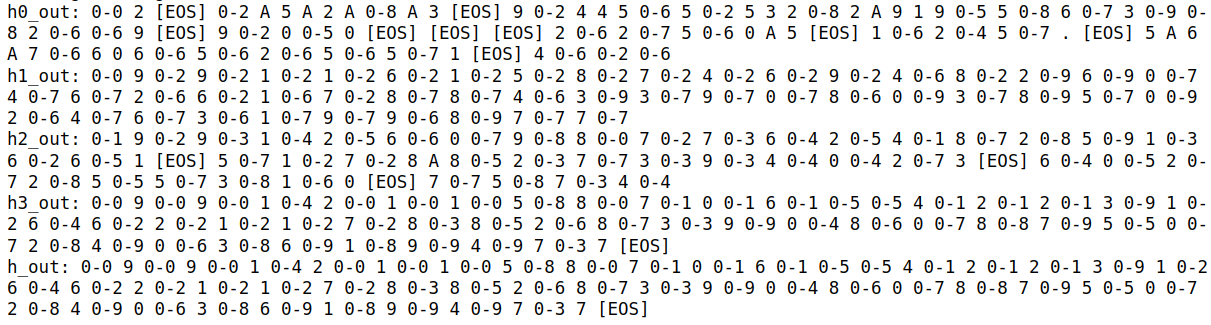}
    \caption{Greedy Decoding with Rank-2 Tokens}
    \label{fig:rank2}
\end{figure}

\begin{figure}[H]
    \centering
    \includegraphics[width=\textwidth]{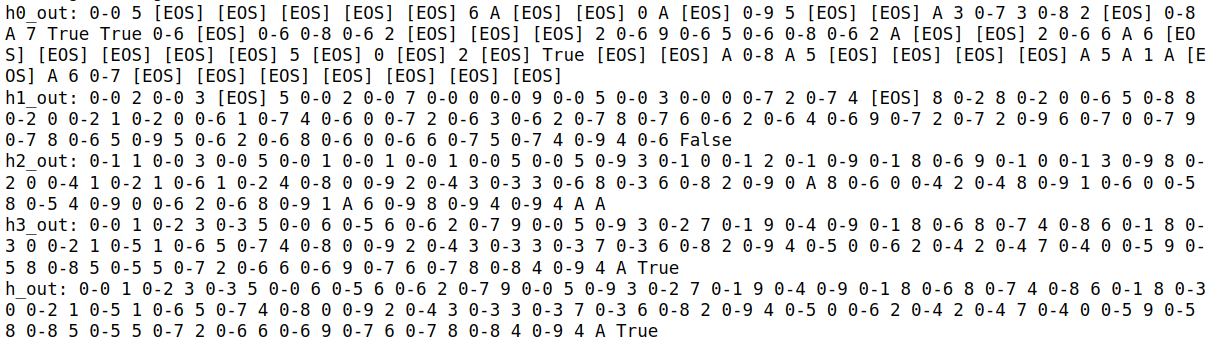}
    \caption{Our Method: Greedy Decoding with Filler Tokens Replaced by Rank-2 Tokens (Recovering hidden characters)}
    \label{fig:our-method}
\end{figure}

\begin{figure}[H]
    \centering
    \includegraphics[width=\textwidth]{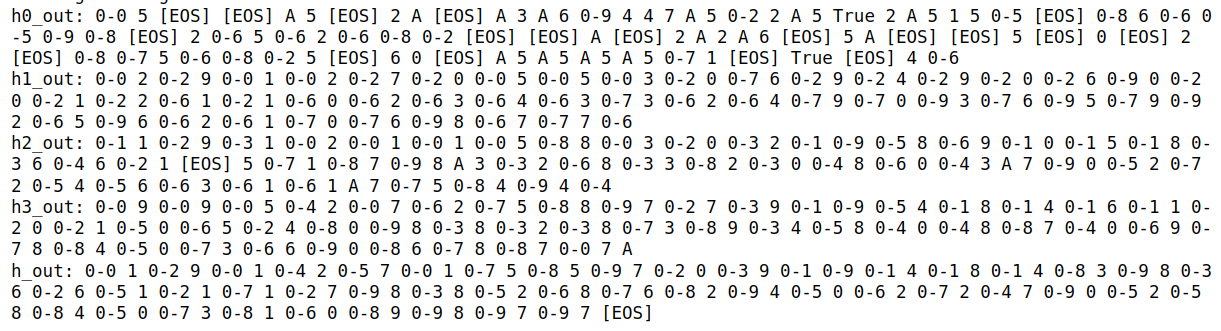}
    \caption{Random Token Replacement: Replacing Filler Tokens with Random Tokens}
    \label{fig:random}
\end{figure}

As shown in Figures \ref{fig:our-method} and \ref{fig:random}, our method successfully recovers the hidden characters, while random token replacement leads to incoherent sequences.

\end{document}